\PassOptionsToPackage{bookmarks={false},hidelinks}{hyperref}
\documentclass[conference,10pt]{IEEEtran}

\usepackage{cite}
\usepackage{amsmath,amsfonts,amsthm}
\usepackage[ruled]{algorithm}
\usepackage[noend]{algpseudocode}
\usepackage[normalem]{ulem}
\usepackage{graphicx}
\usepackage{textcomp}
\usepackage{xcolor}
\usepackage{color}
\usepackage{etoolbox}
\usepackage{subcaption}
\usepackage{multirow}
\usepackage{csquotes}

\usepackage{amsmath,amsfonts,bm}

\def\eqref#1{equation~\ref{#1}}
\def\1{\bm{1}}

\usepackage{hyperref}
\usepackage{url}
\usepackage[english]{babel}

\usepackage{booktabs}
\def\BibTeX{{\rm B\kern-.05em{\sc i\kern-.025em b}\kern-.08em
    T\kern-.1667em\lower.7ex\hbox{E}\kern-.125emX}}

\newbool{inccomment}
\setbool{inccomment}{true}

\newcommand\blfootnote[1]{%
  \begingroup
  \renewcommand\thefootnote{}\footnote{#1}%
  \addtocounter{footnote}{-1}%
  \endgroup
}

\begin{document}
\title{\huge{HCE: Improving Performance and Efficiency with Heterogeneously Compressed Neural Network Ensemble}}

\author{\IEEEauthorblockN{Jingchi Zhang}
\IEEEauthorblockA{\textit{Duke University} \\
Durham, NC, USA \\
jingchi.zhang@duke.edu}
\and
\IEEEauthorblockN{Huanrui Yang}
\IEEEauthorblockA{\textit{University of California} \\
Berkeley, CA, USA \\
huanrui@berkeley.edu}
\and
\IEEEauthorblockN{Hai Li}
\IEEEauthorblockA{\textit{Duke University} \\
Durham, NC, USA \\
hai.li@duke.edu}
}
\maketitle

\begin{abstract}
%As the model sizes of neural networks increasing dramatically, model compression techniques have been widely used for Deep Neural Networks (DNNs). Generally it can be categorized into network pruning that zeros out individual or blocks of weights, and model quantization that converts floating-point parameters to low-precision fixed-point data representations. However, both methods suffer from significant accuracy degradation when compression ratio is ultra high. Meanwhile, ensemble is often used to improve the overall performance of several sub-models with diverse outputs. Our work proposes a new perspective: Can we ensemble highly compressed but heterogeneous sub-models to maintain good performance while keep a low model size and computational cost?
%To answer this question, we introduce a novel training scheme HCE, that generates highly compressed and heterogeneous sub-models by making training procedure aware of the diverse ensemble training object. We also propose a novel visualization method to compare diversities between different DNN models and prove our novel training procedure can introduce diversities in compressed sub-models. We applied our method on both unstructured and structured pruning instances and results show that HCE achieves better performance than its counterpart under the same model size. Furthermore, our method has no limitations on pruning or quantization method and can be easily extend to any model compression method.
Ensemble learning has gain attention in resent deep learning research as a way to further boost the accuracy and generalizability of deep neural network (DNN) models. Recent ensemble training method explores different training algorithms or settings on multiple sub-models with the same model architecture, which lead to significant burden on memory and computation cost of the ensemble model. Meanwhile, the heurtsically induced diversity may not lead to significant performance gain. We propose a new prespective on exploring the intrinsic diversity within a model architecture to build efficient DNN ensemble. We make an intriguing observation that pruning and quantization, while both leading to efficient model architecture at the cost of small accuracy drop, leads to distinct behavior in the decision boundary. To this end, we propose Heterogeneously Compressed Ensemble (HCE), where we build an efficient ensemble with the pruned and quantized variants from a pretrained DNN model. An diversity-aware training objective is proposed to further boost the performance of the HCE ensemble. Experiemnt result shows that HCE achieves significant improvement in the efficiency-accuracy tradeoff comparing to both traditional DNN ensemble training methods and previous model compression methods.
\end{abstract}
\begin{IEEEkeywords}
Ensemble, model compression, efficient neural network
\end{IEEEkeywords}

\blfootnote{Work in progress}
\section{Introduction}

% Ensemble improve performance with diversity -> previous work explore diversity with training same model architecture with different training method -> the design of training method is based on heurstic, repeated model architecture brought redundancy -> can we build efficient ensemble while exploring intrinsic diversity within a model?

% Previous exploration in model compression unveils diverse method to induce efficiency, notably pruning and quantization -> we observe quantization and pruning lead to diverse outcome in decision boundary: quantization retains margin in decision reagion but lead to unsmooth decision boundary, pruning retains smoothness but has smaller margin and worse generalizability -> The intrinsic diversity of pruning and quantization can lead to a diverse ensemble with efficient architecture -> we further propose training objective to boost ensemble performance

Ensemble learning has been widely used to improve the generalization performance of machine learning algorithms by combining multiple diverse weak classifiers~\cite{hansen1990neural,breiman1996bagging,dietterich2000ensemble}. 
%Recently a few studies use ensemble on DNNs models to increase generalization ability~\cite{sinha2021dibs,wen2019batchensemble}. 
An ensemble achieves better accuracy when its sub-models are diverse (i.e. make different errors). Previous deep ensemble methods induce diversity by training multiple DNN models with the same model architecture but different weight initialization and training data~\cite{han2016incremental,huang2018learning,breiman1996bagging,brown2005diversity}.
%However, there is no firm consensus on how to foster the diversity effectively.
However, as the performance and size of DNN models increase dramatically, heuristically ensemble repeated model architecture brought limited performance improvement but heavy computation redundancy. 
This raises a intriguing question: 

\begin{displayquote}
    \textit{Can we build efficient and accurate DNN ensemble while exploring the intrinsic diversity within a pretrained model?}
\end{displayquote}

Previous exploration in model compression unveils diverse methods to induce efficiency, notably pruning and quantization. Quantization aims to convert floating-point parameters into low precision fixed-point representation~\cite{dong2020hawq,yang2020bsq}, whereas pruning aims to remove redundant parameters or structures that have little impact on the accuracy from the model architecture~\cite{wen2016learning,molchanov2019importance}. 
Both methods can lead to significant reduction in model size and computation cost with a small accuracy drop.
In this paper, we make an inspiring discovery that starting from the same pretrained model, quantization and pruning lead to diverse outcome in the decision region: Quantization largely retains (even increase) the margin in decision region, but lead to unsmooth decision boundary due to the discrete representation; Pruning on the other hand retains smoothness of the boundary, but has smaller margin and worse generalizability due to the removal of learnt features. The intrinsic diversity of pruning and quantization can lead to a diverse ensemble with efficient architecture.

Inspired by the analysis above, we propose Heterogeneously Compressed Ensemble (HCE) to improve both performance and efficiency of DNN ensembles. 
We construct HCE as an ensemble of pruned and quantized variants of a pretrained DNN model, which enables us to utilize the intrinsic diversity brought by the heterogeneous model compression methods. Furthermore, we propose a novel diversity-aware model compression method to enable further diversity boost between the compressed variants, therefore improving ensemble performance.
Experiment results show that by assembling specially trained sub-models, higher accuracy can be achieved while maintain low model size and computational cost. 

\begin{figure}
  \centering
  \centerline{\includegraphics[width=8.5cm]{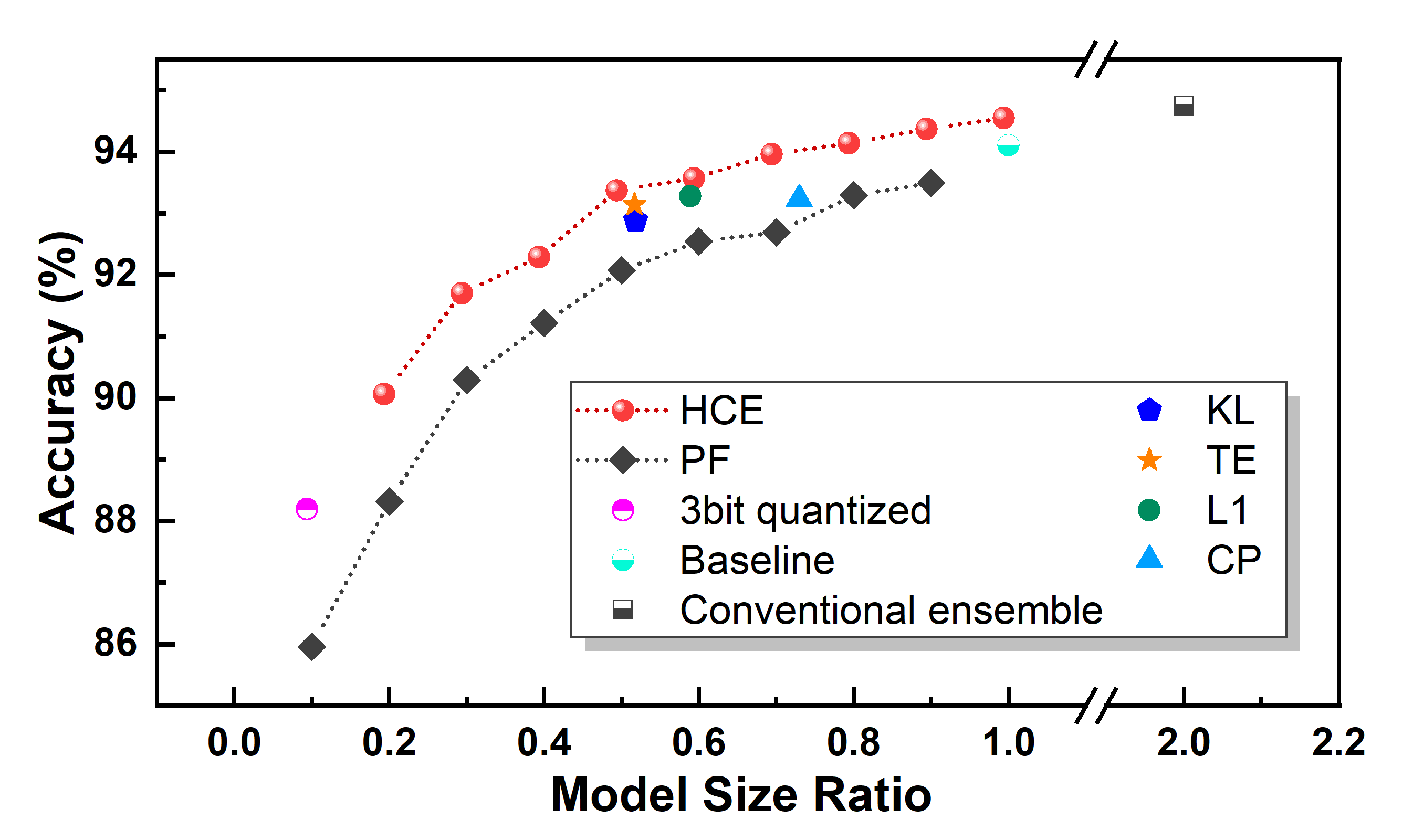}}
  \caption{Comparison of HCE with baselines and other model compression methods on ResNet56, CIFAR-10. Model size ratio denotes the size of the pruned model compared to the 32bit ResNet56. Size of HCE is calculated by the sum of all sub-models in ensemble.}
  \label{fig:comparison}
\end{figure}

Figure~\ref{fig:comparison} shows the result of HCE compared with other SOTA structure pruning methods, including traditional filter pruning~\cite{li2016pruning} (PF), channel pruning~\cite{he2017channel} (CP), L1 norm based method (L1), Taylor expansion~\cite{molchanov2019importance} (TE) and KL-divergence metric~\cite{luo2020neural} (KL). In this case, HCE uses uniform quantization and filter pruning~\cite{li2016pruning}. The model size of HCE is the sum of both compressed sub-models. As shown, HCE achieves better accuracy than the original baseline model and other SOTA structural pruning methods under the same model size. We also include a conventional deep ensemble method that trains two full ResNet56 with different initialization. Result shows that HCE achieves comparable accuracy with half of the size.

To the best of our knowledge, HCE is the first work to build deep ensemble with heterogeneous model compression techniques. We make following contributions:
\begin{itemize}
    \item We propose HCE, a novel efficient DNN ensemble training method that utilizes and enhances the intrinsic diversity between heterogeneously compressed sub-models.
    \item We unveil the intrinsic difference between pruned and quantized model via the analysis of decision regions.
    \item We propose a diversity-aware knowledge distillation objective for compressed sub-model that will enable higher pruning rate and diversity in sub-model. 
    \item We perform extensive experiment on CIFAR-10 and ImageNet, where we show HCE achieves better efficiency-accuracy tradeoff than SOTA model compression methods and deep ensemble methods.
\end{itemize}

The rest of the paper is organized as follows. We first provide background and related work on ensemble and model compression in Sec.~\ref{sec:related}. Then we introduce our decision region visualization and analysis method and training procedure for HCE in Sec.~\ref{sec:method}. In Sec.~\ref{sec:experiment} we discuss the experiment result and show effectiveness of HCE.
\section{Related work}
\label{sec:related}

%Recently a few studies use ensemble on DNNs models to increase generalization ability~\cite{sinha2021dibs,wen2019batchensemble}. 

\subsection{Ensemble}

Ensemble learning has been widely used to improve the performance of machine learning models.
%Currently DNN models show better performance in most machine learning applications compared to traditional techniques like regressions and SVMs. 
By combining the good performance of DNN models and the improved generalization ability from ensemble method, deep ensemble learning models generally show better performance compared to both traditional ensemble models and single DNN model. 
%However, current deep ensemble methods are naive and do not fully utilize the potential of deep learning and ensemble. 

Ensemble models can be broadly categorised into homogeneous and heterogeneous ensemble, based on if each sub-model within the ensemble takes the same model architecture or not. 
%Generally, homogeneous ensemble has the same algorithm for each sub-model but trained with distinct datasets. As for heterogeneous ensemble, sub-models are trained with same dataset but using different algorithms. Hence each sub-model can be as diverse as possible.
In homogeneous ensembles, diversity is induced in the training process since all the model architectures are identical. 
%For example, sampling the feature space of the training data and train multiple models with different subsets of the training set.
For neural networks, training multiple sub-models with different initialization is proved to be an effective way to induce diversity~\cite{han2016incremental,huang2018learning}. Also, some works use specially designed diversity training objective to increase robustness against adversarial attacks~\cite{yang2020dverge}. However, current DNN models have large model size and high training and inference costs, so that training multiple independent models is an inefficient option~\cite{laine2016temporal}. Some works have been done to solve this problem ~\cite{xie2013horizontal} but it is still challenging to train multiple DNN sub-models for ensemble since too many parameters and hyper-parameters need to be tuned and optimized. 

On the other hand, heterogeneous ensemble methods show the edge of lower computational cost and higher diversity. Some studies used DNN models combined with traditional models, such as support vector machine (SVM) or random forest (RF), to lower computation and increase diversities~\cite{kilimci2018deep}. By using different data subsets, model architectures and fusion strategy, heterogeneous ensemble will introduce more diversity and hence show better generalization performance.
%Current heterogeneous ensemble shows good effectiveness on applications such as image recognition and text classification~\cite{kilimci2018deep}. 
%By combining deep learning models with conventional models such as support vector machine (SVM), random forest (RF), logistic regression, etc, heterogeneous ensemble showed more diverse and hence provide better generalization performance. 

Our proposed method tries to take advantage of both homogeneous and heterogeneous ensemble. We utilize multiple DNN models for their good performance as in homogeneous ensemble, meanwhile introducing intrinsic model architecture diversity in sub-models via heterogeneous compression. Also, we keep sub-models in small size while introduce diverse model architecture for better ensemble generalization ability.

\subsection{Model Compression}
% Both methods can reduce the model size and computational cost but suffer from severe accuracy degradation when compression rate is high.

Model compression for DNNs have been widely used to accelerate models on resource-limited devices. Two popular methods are pruning and quantization.

Neural network pruning aims to zero out redundant parameters to reduce both model size and computation cost. Generally it can be categorized into 1) unstructured pruning and 2) structured pruning. Unstructured pruning can achieve high sparse ration with negligible accuracy drop~\cite{lee2020layer,liu2018rethinking,mocanu2018scalable,frankle2018lottery,cai2020rethinking,yang2019deephoyer}. However, unstructured pruning can hardly bring actual speedup on hardware because of the irregular non-zero data structure.  On the other hand, coarse-gained structured pruning can benefit real-life hardware inference such as filter pruning~\cite{wen2016learning} and channel pruning\cite{li2016pruning}. Although structured pruning can generate hardware-friendly weight structures, it cannot achieve relatively high sparse ratio compared to unstructured pruning.
%Recently some new pruning methods come up due to the support of new hardware such as N:M pruning\cite{mishra2021accelerating} that apply fine-grained sparsity and generate hardware-friendly data structure with specific hardware design.

Quantization is another common and effective way to compress the deep learning models. It reduces the DNN model size and lower computation cost by replacing the floating point weights with low precision fixed-point data. Common quantization methods including directly apply uniform quantizers~\cite{morgan1991experimental,wu2016quantized}, quantization-aware fine-tuning~\cite{zhou2017incremental} and mixed-precision quantization~\cite{yu2022hessian,dong2020hawq,yang2020bsq}. Quantization can largely decrease DNN's arithmetic intensity but still cause significant accuracy degradation for ultra-low data precision.

Our work explores the intrinsic diversity in the compressed model decision boundary induced by different compression methods. HCE utilizes the diversity between pruned and quantized model to build efficient ensemble, and further improves the ensemble performance with a diversity-aware sub-model training objective.

\section{Method}
\label{sec:method}
In this section, we will first analysis the diversity of heterogeneously compressed DNN models by visualizing the decision region. Then we propose the HCE architecture and training scheme, which is an efficient way to produce highly-compressed sub-models for deep ensemble while guarantee high diversity within each individual sub-model. We visualize the improvement on sub-model diversity and ensemble decision regions brought by HCE in the end.

\subsection{Decision Region Visualization}

\begin{figure}[tb]
\begin{minipage}[b]{0.45\linewidth}
  \centering
  \centerline{\includegraphics[width=4.5cm]{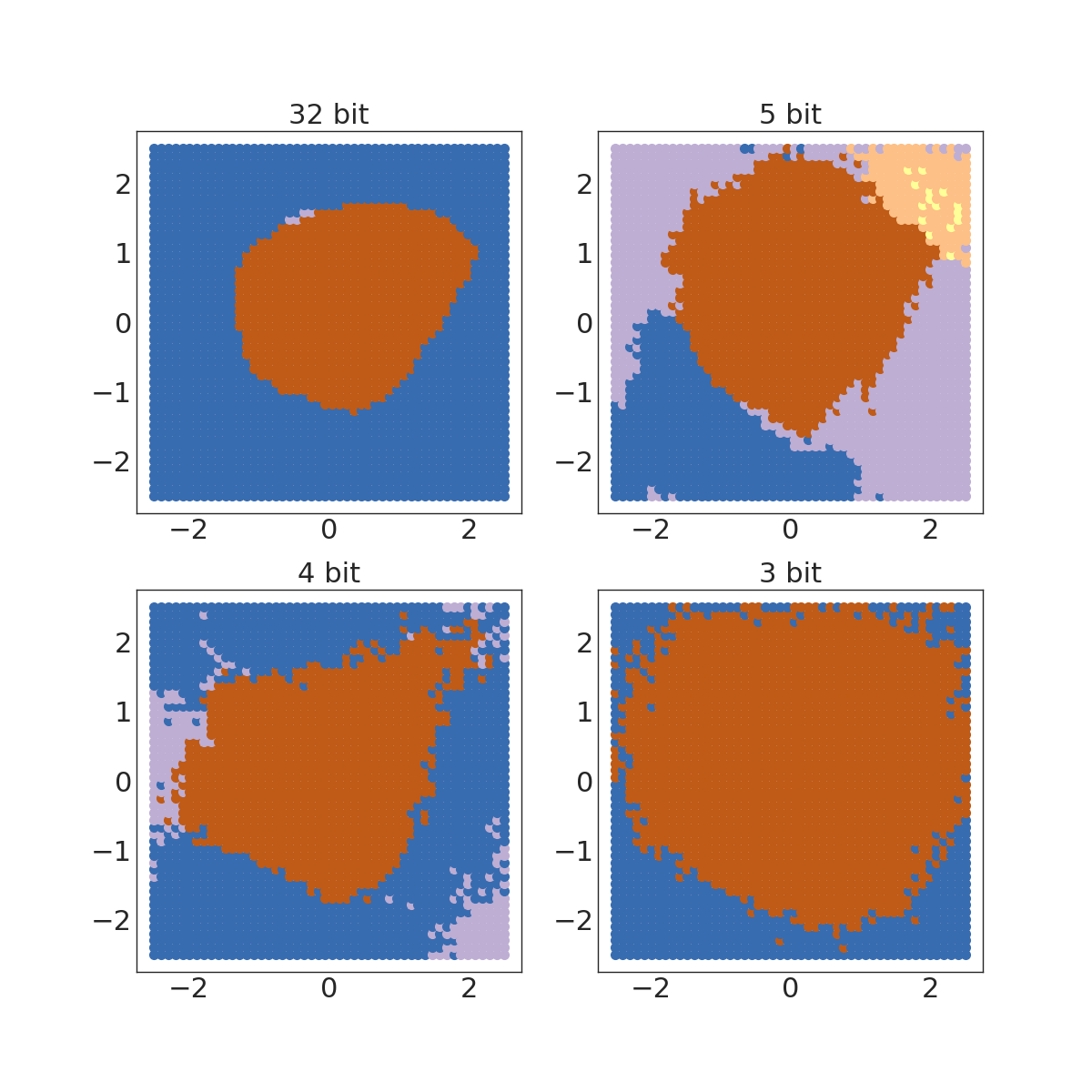}}
%  \vspace{1.5cm}
  \centerline{(a) {Quantized}}
\end{minipage}
\hfill
\begin{minipage}[b]{0.45\linewidth}
  \centering
  \centerline{\includegraphics[width=4.5cm]{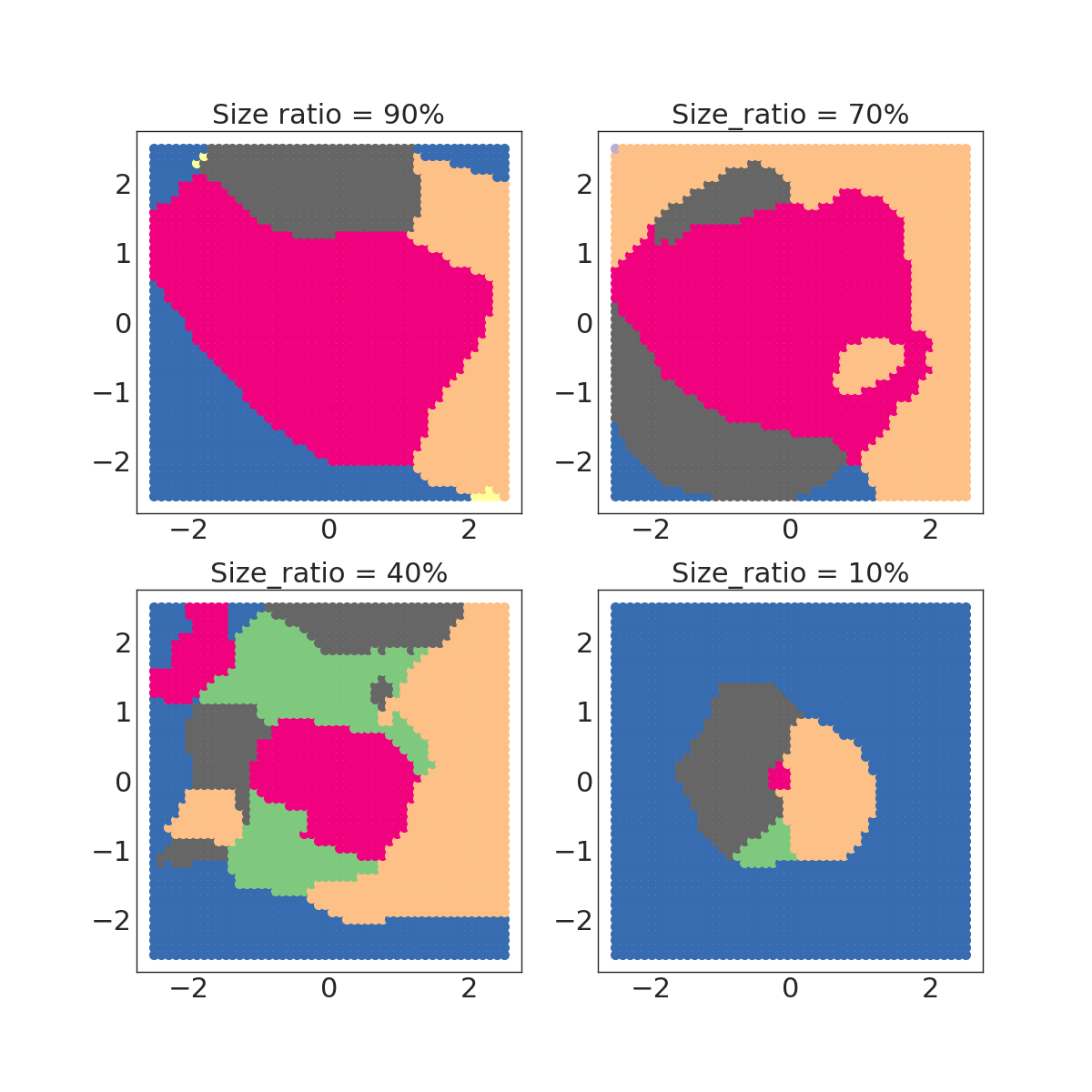}}
  \centerline{(b) {Pruned}}
\end{minipage}
\caption{Decision region visualizations of quantized and pruned model of a ResNet56 trained on CIFAR-10. The vertical and horizontal axes are two random Rademacher vectors. The zero point represents a test data point. Same color indicates the predicated label. Decision region can be inferred from the shape of the pattern with the same color.}
\label{fig:DBV}
\end{figure}

We first illustrate how heterogeneously compressed DNN models learn different representations in the feature space. Figure~\ref{fig:DBV} shows how decision region changes in the quantized and pruned models when the quantization precision and the sparsity varies.
We first trained a 32-bit ResNet56 model on CIFAR-10 as baseline. For Figure~\ref{fig:DBV}(a), we generate a series of quantized models, 5-bit, 4-bit and 3-bit, progressively using post-training uniform quantization method. Also for Figure~\ref{fig:DBV}(b), We generate a series of structure pruned models with 90\%, 70\%, 40\% and 10\% size ratio using filter pruning~\cite{li2016pruning}. No finetuning is performed after the compression.
For all the subplots in each figure, we plot the decision boundaries in the vicinity of the same testing data point and with the same scale on a 2-D plane span by two random vectors.

Figure~\ref{fig:DBV}(a) shows that when bit precision decreases, the decision region grows larger correspondingly while the decision boundaries become unsmooth. In detail, the top left figure in \ref{fig:DBV}(a) is the decision region of the pre-trained baseline 32-bit ResNet56 model. When quantize the model into 5-bit, the decision region enlarges and the surface gets larger and rougher compared to the baseline. This trend follows when bit precision further decreases to 4-bit and 3-bit. One explanation for this decision region change is, since we quantize both weight and activation parameters in the model, the output feature of each layer becomes inaccurate and sensitive to input disturbs. The quantized activation makes the output feature more discrete so that the decision region may vary a lot when there is small perturbation on input. In fact, the decision region does not expand after quantized but get sensitive to input disturbs. However, with quantized activation layers, the output feature falls into discrete spaces. That is why the decision boundary gets unsmooth when data precision decrease.

Unlike the quantized, the pruned case shows the opposite phenomenon. Figure~\ref{fig:DBV}(b) shows the trend of decision region when the sparisty increases in the model. In this case, we use structure pruning method to prune the filter inside the baseline model. As is shown in Figure~\ref{fig:DBV}(b), when sparsity level increases, i.e. size ratio decreases, the decision boundaries get smaller. This is because when structure pruning zeros out filters in the model, fewer features can be extracted to feature space. With fewer information extracted, the generalization ability of the DNN model decreases and so as the margin of decision boundary.

From above, we can infer that although compressed from the same baseline model, quantized and pruned models show great difference on feature representation in hyperspace. From generalization perspective, quantization and pruning both effect the sensitiveness of the DNN model. The difference is that quantization makes the decision boundary unsmooth and sensitive to perturbation and pruning decrease the generalization capability but the decision boundary remain smooth. This difference introduces different properties in quantization and pruned models. By applying different model compression methods, we can foster diversity in deep ensembles.

\subsection{Training Method}
\label{sec:training}

\begin{figure}
  \centering
  \centerline{\includegraphics[width=8cm]{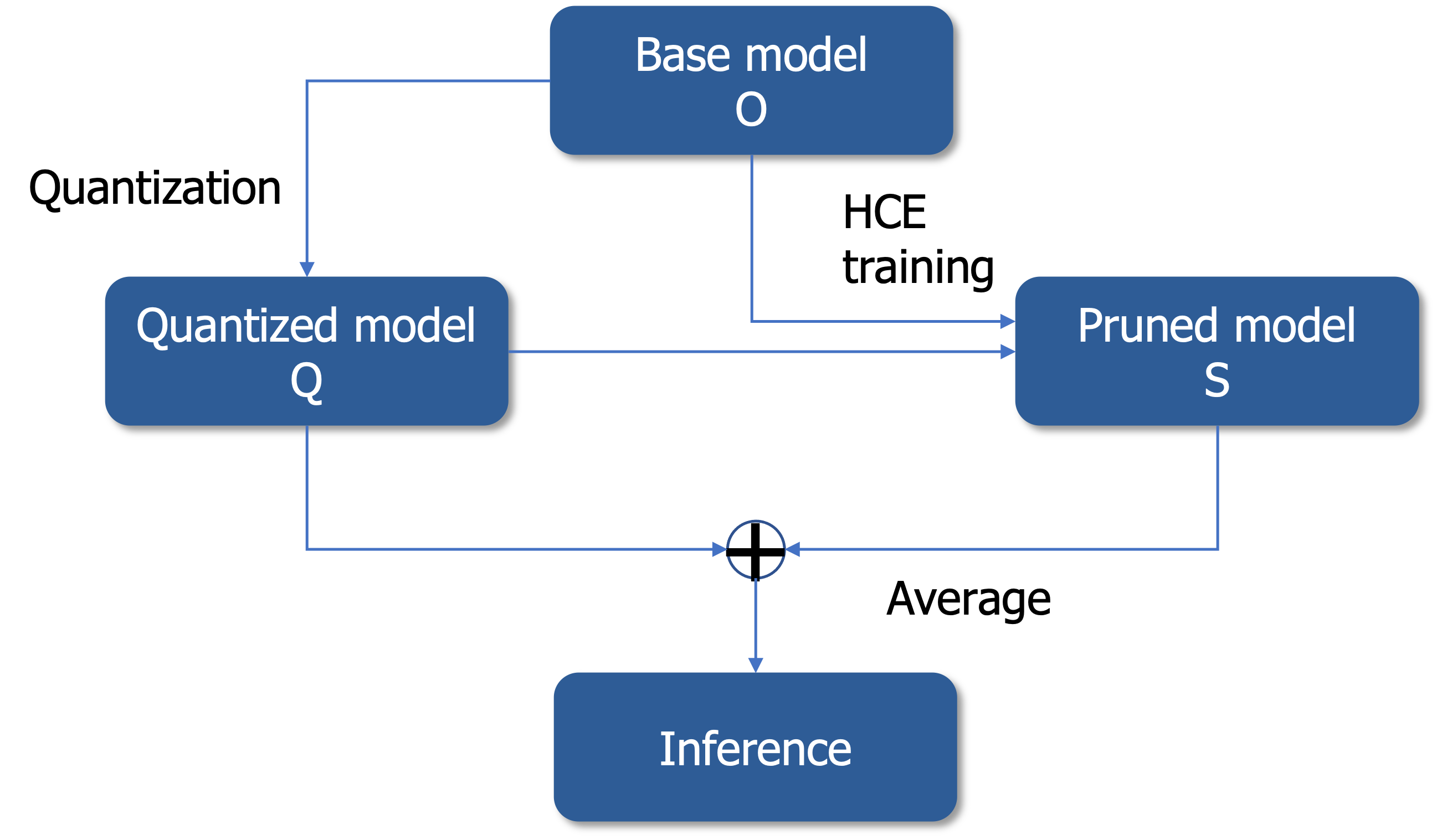}}
  \caption{training pipeline of HCE}
  \label{fig:training_step}
\end{figure}

Based on previous discussion, we propose our novel HCE training scheme, that ensembles the quantized and pruned model to introduce more diversity while maintaining low computation cost.
Our HCE training method is composed of four steps. As is shown in Figure \ref{fig:training_step}, the HCE training algorithm is designed as follows: 1) Train the full-precision, full-size model O as baseline. 2) Quantize the baseline model and get low-precision model Q. 3) Iteratively prune a sparse model based on O and Q while training with our proposed ensemble-aware training object. 4) For inference, simply average the output of quantized model and pruned model.
Here we propose to fix the quantized model while training the pruned model for accuracy improvement, because the discrete weight and activation representation in the quantized model may hinder the stability of the training process.

Specifically, the pruned model is trained with the object function in Equation~(\ref{equ:loss}). 

\begin{equation}
\label{equ:loss}
\begin{split}
&Loss(S)= \alpha*L_{CE}(output(S)) +\\
&(1-\alpha)*L_{KL}(output(S), (output(O)- output(Q)) \\
\end{split}
\end{equation}

The object function is inspired by Knowledge Distillation (KD) training. The first part is the cross entropy of the sparse model S. The latter part of the object function is the KL divergence between the output probability of the base model O and the quantized model Q. This part keeps the pruned model focus on the information that lost during the quantization. $\alpha$ is a coefficient that controls the level of learning from the difference and directly from the training data.
In detail, given a input $x$, the base model would produce a score vector $s^O(x)=[s_1^O(x),s_2^O(x),...,s_k^O(x)]$ and convert to probability by softmax: $p_k^O(x)=\frac{e^{s_k^O(x)}}{\sum_je^{s_j^O(x)}}$. We also "soften" the probabilities using temperature scaling~\cite{guo2017calibration}:
\begin{equation}
\label{equ:prob}
p_k^O(x)=\frac{e^{s_k^O(x)/\tau}}{\sum_je^{s_j^O(x)/\tau}}
\end{equation}
where $\tau>1$ is the temperature hyperparameter. 
Same as base model, quantized also generates a corresponding probability: $p_k^Q(x)=\frac{e^{s_k^Q(x)/\tau}}{\sum_je^{s_j^Q(x)/\tau}}$ and training sparse model has its own prediction $p_k^S(x)$. So the probability difference is $p_k^D(x)=p_k^O(x)-p_k^Q(x)$ and the KL divergence loss is:
\begin{equation}
\label{equ:KL}
L_{KL}=-\tau ^2\sum_k p_k^D(x)\log P_k^S(x)
\end{equation}

Unlike training a student model from the teacher model in knowledge distillation, we try to force the pruned model to focus more on the prediction gap between the base model O and quantized model Q. Since quantized models often remain acceptable accuracy so that the gap between O and Q is limited. With less information to learn, the pruned model can achieve higher pruning rate.

The goal of our training objective in Equation~(\ref{equ:loss}) is to make the output of the ensembled models close to that of the original model. In other word, the pruning method is aware of the diverse ensemble training. The KL divergence push the sparse model to produce output logits $output(S)$ similar to $output(O)-output(Q)$. In this case, during the inference when ensembling the pruned model $S$ and quantized model $Q$, the output $output(S)+output(Q)$ should be close to the output of the original model  $output(O)$. Our experiment results prove that performing iterative pruning with the proposed diversity-aware training objective can lead to very sparse sub-model that can still result in high ensemble accuracy.

It is worth to point out that our method has no limitations on the quantization or pruning methods. Our experiments will demonstrate that both the structural or unstructural pruning can be effective.

\subsection{HCE Diversity Analysis}

\begin{figure}
  \centering
  \centerline{\includegraphics[width=7cm]{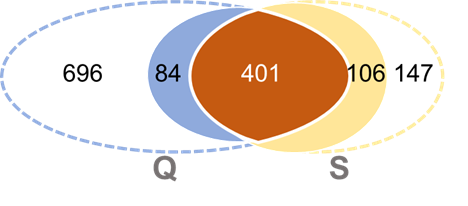}}
  \caption{Venn figure of wrongly predicted images of ResNet56 on CIFAR-10. Numbers in the dash circles indicates the image wrongly classified by individual sub-model but corrected by the ensemble.}
  \label{fig:contour}
\end{figure}

\begin{figure}[tb]
\begin{minipage}[b]{0.45\linewidth}
  \centering
  \centerline{\includegraphics[width=4.5cm]{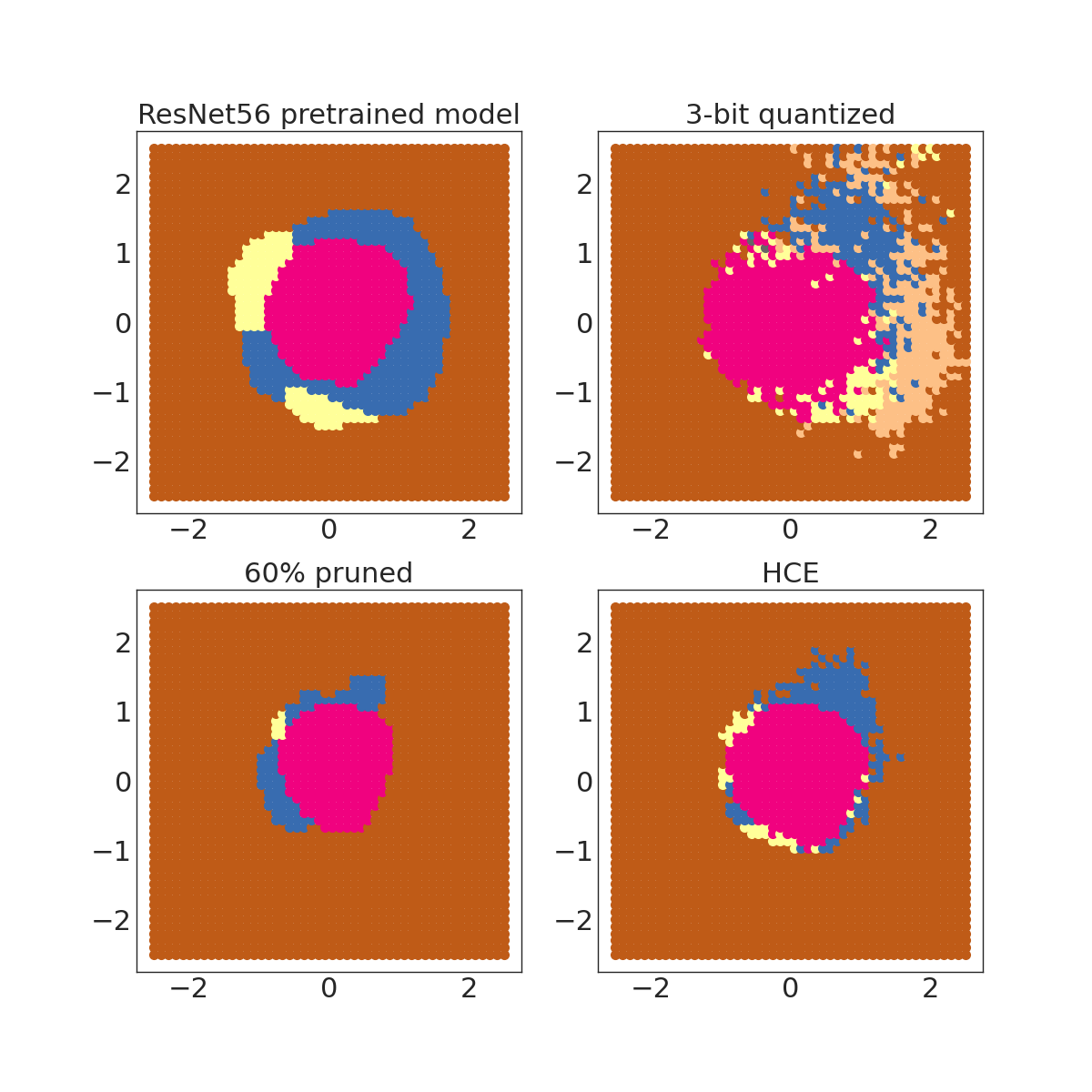}}
%  \vspace{1.5cm}
  \centerline{(a)}
\end{minipage}
\hfill
\begin{minipage}[b]{0.45\linewidth}
  \centering
  \centerline{\includegraphics[width=4.5cm]{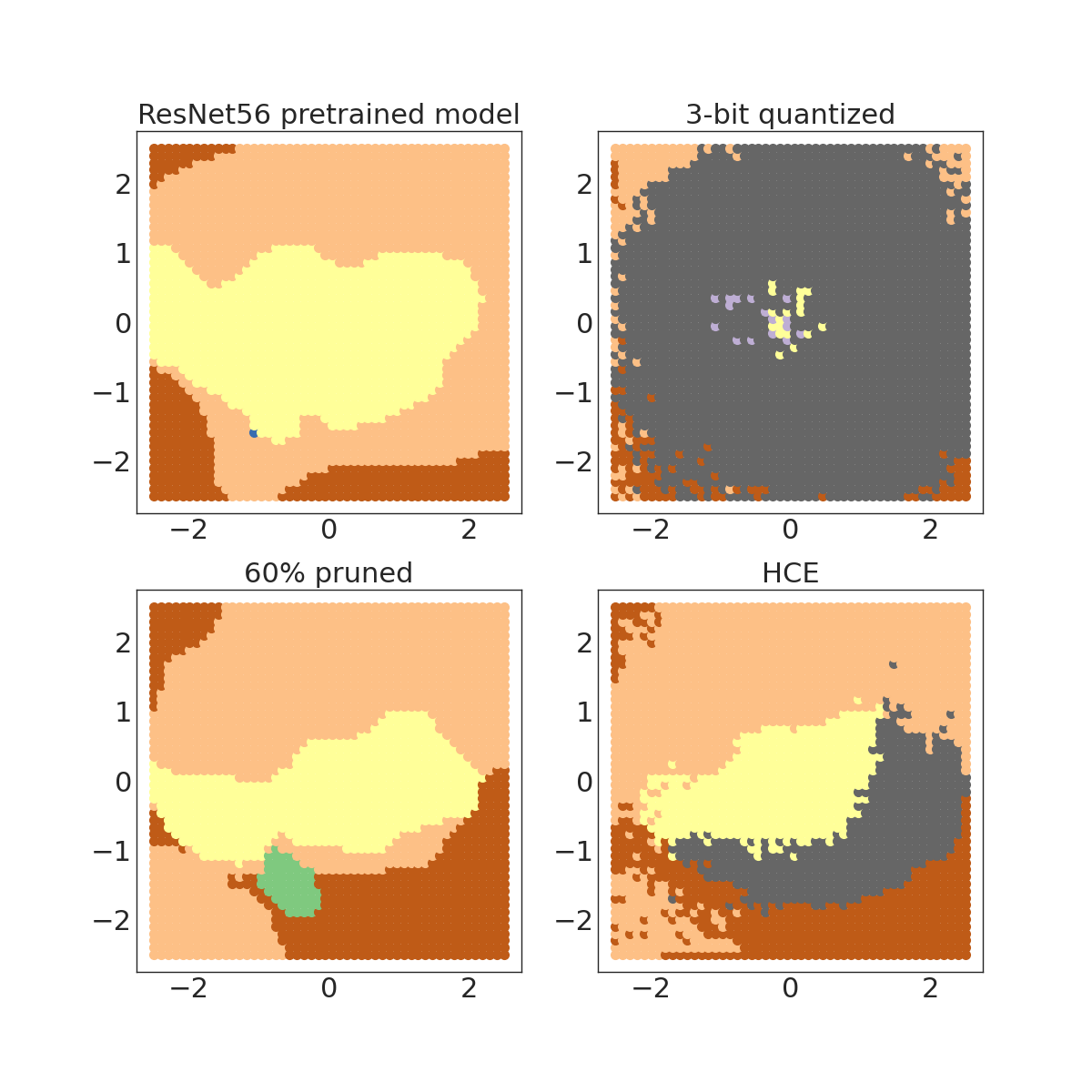}}
  \centerline{(b)}
\end{minipage}
\caption{Decision region visualizations showing (a) Quantized and HCE both make correct prediction and (b) HCE corrects mispredicted labels from quantized model}
\label{fig:QSE_wrong}
\end{figure}

To further analyse the effectiveness of our ensemble method, we look into the wrongly predicted labels of each individual sub-model. Still, We evaluate our method on ResNet56 model on the CIFAR-10 dataset. We first generated a 3-bit quantized model from the pretrained model and then trained a 60\% pruned model filter pruning~\cite{li2016pruning} with the HCE objective. As Shown in Figure~\ref{fig:contour}, among 10,000 testcases, the baseline model have 590 error cases. The quantized submodel has 1,181 error cases and HCE pruned submodel has 654 error cases. However, among all error cases, only 401 cases are overlapped. The percentage of overlapped error cases in all error cases is $(E_Q\cup E_S) / (E_Q\cap E_S) = 27.96\%$, which denotes that although quantized and pruned model are trained from the same pretrained model, they make very different error predictions, i.e. diversity. After ensemble, HCE produces similar accuracy as the baseline. 58.79\% errors made by individual sub-model are corrected, including 90\% of the errors made by quantized model alone. This indicates the proposed training objective is effective.

We also compare the decision boundaries of HCE under different situations in Figure~\ref{fig:QSE_wrong}. Figure~\ref{fig:QSE_wrong}(a) shows that, as we discussion above, the decision boundary of quantized model is unsmooth while that of pruned model has small margin. HCE can restore the behavior of the original model by ensembling two compressed model. Figure~\ref{fig:QSE_wrong}(b) shows the case where quantized model makes mistake while HCE corrects the predictions. As the quantized model is perturbed into a wrong prediciont, the smooth decision region of the pruned model enables the ensemble to achieve correct output. This show that our HCE can indeed introduce and make use of the diverse model compression methods and produce similar performance as the original full-precision model.

\section{Experiment Result}
\label{sec:experiment}
\subsection{Setup}
We compare HCE with various model compression counterparts. The base ResNet is implemented following~\cite{he2016deep} and we used both ResNet20 and ResNet56 on CIFAR-10. On ImageNet we tried HCE with ResNet50 and MobileNetV2. We used conventional uniform quantization and filter pruning~\cite{li2016pruning} for experiment but HCE can be extended on other model compression methods. We also conduct an ablation study on hyper-parameters $\alpha$ mentioned in Section~\ref{sec:training}. To make comparable estimation on the computation cost of quantized and sparse model, we use a unified notion of FLOPs. Since quantized model only requires fixed-point operations, we first calculated the bit operations (BOPs) for the quantized model, then convert it to an estimated FLOPs by dividing a scaling factor of 23, which is the number of fraction bits in a Float32 representation~\cite{ron2022experimental}.

\subsection{CIFAR-10 Result}
\label{cifar10_result}
We first evaluate HCE on CIFAR-10 datasets and results are shown in Table~\ref{tab:cifar10_structure}. The FLOPs of HCE is measured for both compressed sub-models in HCE. The quantized model in HCE is 3bit with accuracy of 88.19\%. For large models like ResNet56, HCE can remove more than half FLOPs and achieve 93.56\% accuracy, higher than other SOTA structured pruning methods. One important information is that, if we directly apply the PF method, the accuracy is 92.07\% when FLOPs is 50\%. But in our experiment, our pruning method is also PF combining with our training objective. By ensembling with a 3-bit model, the accuracy increases 1.5\% with less FLOPs. The pruning method in HCE can be substituted with any other structured pruning method like DCF and SFP and HCE result can be better.

As for smaller model ResNet20, HCE still achieved best accuracy compared to all other counterparts. With 57\% FLOPs, the ensemble accuracy can achieve 91.99\%. The hyperparameter $\alpha$ in training object is 0.2 and 0.4 individually.

%The results show that for both ResNet56 adn ResNet20, HCE achieve higher accuracy than other structured pruning method under similar FLOPs ration.

\begin{table}[tb]
\caption{HCE results on CIFAR-10  }
\label{tab:cifar10_structure}
\centering
%\setlength{\tabcolsep}{2.5pt}
%\begin{tabular}{p{1.24cm}p{0.6cm}p{1.08cm}p{0.9cm}p{0.9cm}p{1.3cm}}
\vspace{3pt}
\begin{tabular}{cccc}
\toprule
Model & Approach & Accuracy & FLOPs(M) \\

\midrule

%& & Acc(\%) & Acc(\%) & Acc Drop (\%)& Flop Reduction(\%)  \\
 ResNet56 & Baseline                                 & 93.80  & 126.8 (100\%)  \\
          & PF\cite{li2016pruning}                   & 92.07  & 63.4 (50.0\%) \\
          & CP\cite{he2017channel}                   & 91.80  & 63.4 (50.0\%) \\
          & AMC\cite{he2018amc}.                     & 91.90  & 63.4 (50.0\%) \\
          & DCP\cite{zhuang2018discrimination}       & 93.49  & 63.4 (50.0\%) \\
          & SFP\cite{he2018soft}                     & 93.35  & 59.6 (47.1\%) \\
          & FPGM\cite{he2019filter}                  & 93.49  & 59.6 (47.1\%) \\
          & CCP\cite{peng2019collaborative}          & 93.46  & 67.2 (53.0\%) \\
          & \textbf{HCE($\alpha$=0.2)}             & \textbf{93.56}  & 60.6 (47.8\%) \\
\midrule
 ResNet20 & Baseline                   & 92.20  & 41.2 (100\%) \\
 & SFP\cite{he2018soft}                & 90.83  & 23.9 (58.0\%)  \\
 & FPGM\cite{he2019filter}             & 91.99  & 19.0 (46.1\%)  \\
 & \textbf{HCE($\alpha$=0.4)}        & \textbf{92.35}  & 23.4 (56.8\%)    \\

\bottomrule
\end{tabular}
\end{table}

\subsection{ImageNet result}
\begin{table}[tb]
\caption{HCE results on ImageNet}
\label{tab:imagenet_structure}
\centering
%\setlength{\tabcolsep}{2.5pt}
%\begin{tabular}{p{1.24cm}p{0.6cm}p{1.08cm}p{0.9cm}p{0.9cm}p{1.3cm}}
\vspace{3pt}
\begin{tabular}{cccc}
\toprule
Model & Approach & Accuracy & FLOPs(G) \\

\midrule

 ResNet50 & Baseline                                 & 76.01  & 4.1 (100\%)  \\
          & DCP\cite{zhuang2018discrimination}       & 74.95  & 1.8 (44.5\%) \\
          & FPGM\cite{he2019filter}.                 & 74.83  & 1.9 (46.5\%) \\
          & AutoPruner\cite{luo2018autopruner}       & 74.76  & 2.0 (48.8\%) \\
          & SFP\cite{he2018soft}                     & 74.61  & 1.7 (41.8\%) \\
          & \textbf{HCE($\alpha$=0.3)}             & \textbf{75.13}  & 1.8 (44.7\%)  \\
\midrule
 MobileNetV2 & Baseline                 & 71.8   & 0.32 (100\%) \\
 & AMC\cite{he2018amc}                  & 70.80 & 0.22 (70.0\%)  \\
 & MetaPruning\cite{liu2019metapruning} & 71.2  & 0.23 (72.3\%) \\
 & \textbf{HCE($\alpha$=0.5)}           & \textbf{71.35}  &  0.23 (71.8\%)   \\

\bottomrule
\end{tabular}
\end{table}

For ImageNet dataset, we test our HCE on ResNet50 and MobileNetV2 and the result is shown in Table~\ref{tab:imagenet_structure}. For ResNet50, HCE achieve highest accuracy compared to other popolar structured pruning methods under similar FLOPs ratio. We also try HCE on MobileNetV2. MobileNetV2 is a reletive small and thin DNN model compared to ResNet50 so compress MobileNetV2 on ImageNet is not easy. We would like to see how HCE works on such small models. As is shown in Table~\ref{tab:imagenet_structure}, HCE also achieves better performance under similar FLOPs. 

However, in ImageNet experiments, $\alpha$ is 0.3 on ResNet50 and 0.5 on MobileNetV2. We will further discuss this $\alpha$ hyperparameter later.

\subsection{Ablation study on $alpha$}
\begin{figure}
  \centering
  \centerline{\includegraphics[width=8cm]{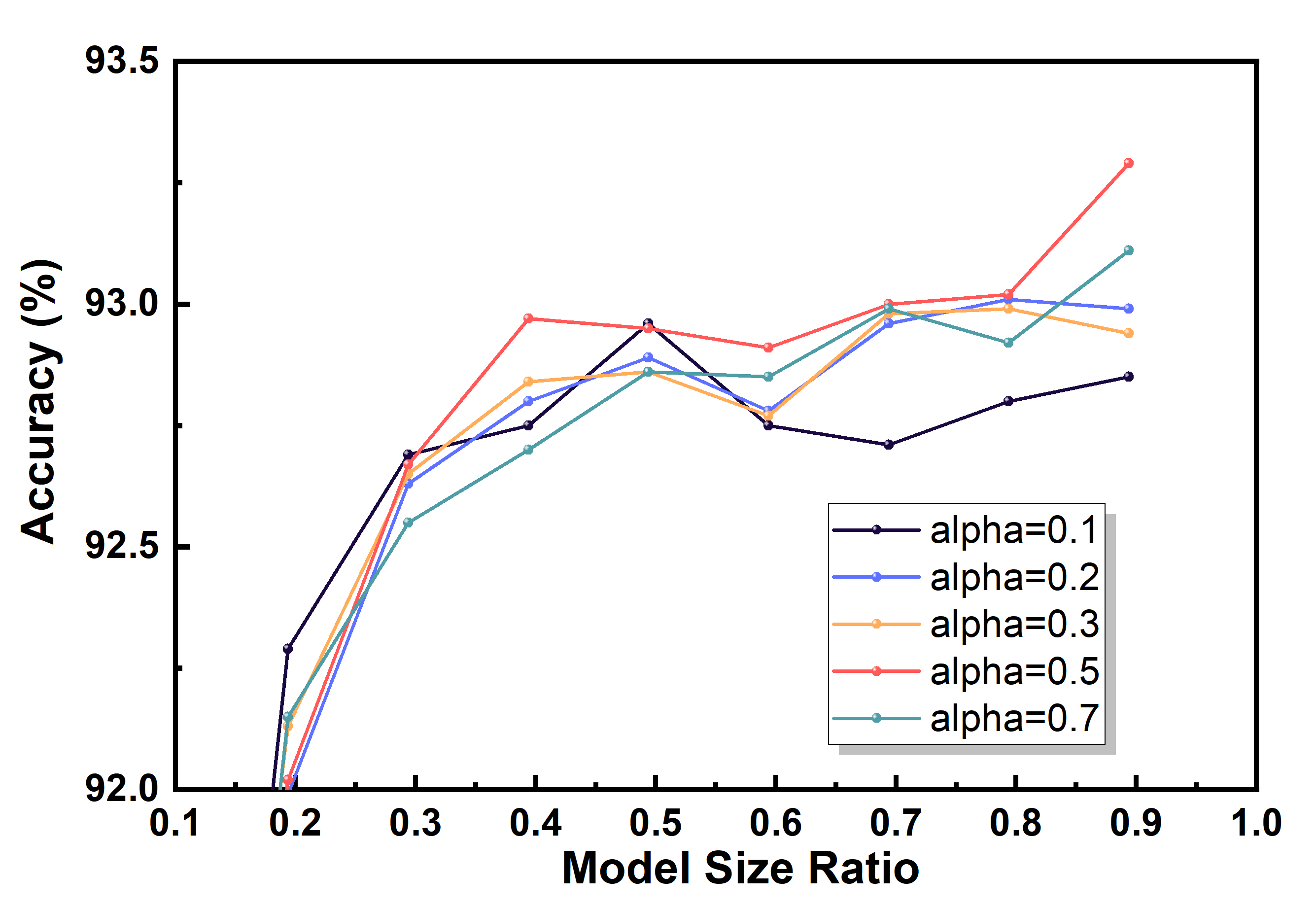}}
  \caption{Ablation study on hyperparameter $\alpha$}
  \label{fig:ablation}
\end{figure}

To study the effectiveness of the hyperparameter $\alpha$ in section\ref{sec:training}, we did a series experiment by tuning $\alpha$. In this ablation study we apply HCE on ResNet20 on CIFAR-10. Different from above experiment setting, we applied L1 unstructured pruning in this case to demonstrate the extension ability of HCE. As is shown on Figure~\ref{fig:ablation}, on ResNet20 unstructured pruning case, $\alpha$=0.5 achieves higher accuracy than other $\alpha$ values in most cases and seems to be superior than other values. Meanwhile $\alpha$ =0.1 always produces lowest accuracy. However, on structured pruning of ResNet56 in section~\ref{cifar10_result}, $\alpha$=0.2 is the best choice. The explanation is that for our training objective~\ref{equ:loss}, $\alpha$ denotes the portion of HCE learning from the cross entropy of the input data while (1-$\alpha$) represents the learning from the difference between baseline and the quantized model. For large models, learning too much from the difference will cause the heavy overfitting and degrade the accuracy. But for small models like MobileNetV2 for ImageNet, overfitting will not occur when $\alpha$ increases.

\section{Conclusion}

%This work proposes HERO, a Hessian-enhanced robust optimization method to improve the generalization and quantization performance of DNN models simultaneously. We provide novel insights on unifying generalization and quantization under improving weight perturbation robustness, theoretical analysis on enhancing the robustness with Hessian regularization, and empirical results showing the effectiveness of HERO. We hope this work helps on deploying DNN models onto real-world mobile and edge devices, and inspires further attention to the robustness against weight perturbation.

This work propose HCE, a novel scheme to build heterogeneous deep ensemble to improve the generalization and efficiency simultaneously. We also introduce a new method to illustrate the intrinsic diversity within compressed models and show that HCE can generate highly compressed but diverse sub-models. Both CIFAR-10 and Imagenet experiments show the effectiveness of HCE. We hope our work can bring new perspective on deep ensemble and help deploying ensembles on resource-limited devices. 

\bibliographystyle{unsrt}
\bibliography{main}

\begin{thebibliography}{10}

\bibitem{hansen1990neural}
Lars~Kai Hansen and Peter Salamon.
\newblock Neural network ensembles.
\newblock {\em IEEE transactions on pattern analysis and machine intelligence},
  12(10):993--1001, 1990.

\bibitem{breiman1996bagging}
Leo Breiman.
\newblock Bagging predictors.
\newblock {\em Machine learning}, 24(2):123--140, 1996.

\bibitem{dietterich2000ensemble}
Thomas~G Dietterich.
\newblock Ensemble methods in machine learning.
\newblock In {\em International workshop on multiple classifier systems}, pages
  1--15. Springer, 2000.

\bibitem{han2016incremental}
Shizhong Han, Zibo Meng, Ahmed-Shehab Khan, and Yan Tong.
\newblock Incremental boosting convolutional neural network for facial action
  unit recognition.
\newblock {\em Advances in NeurIPS}, 29, 2016.

\bibitem{huang2018learning}
Furong Huang, Jordan Ash, John Langford, and Robert Schapire.
\newblock Learning deep resnet blocks sequentially using boosting theory.
\newblock In {\em International Conference on Machine Learning}, pages
  2058--2067. PMLR, 2018.

\bibitem{brown2005diversity}
G~Brown, Jeremy Wyatt, Rachel Harris, and Xin Yao.
\newblock Diversity creation methods: A survey and categorisation.
\newblock {\em Information Fusion}, 6(1):5--20, 2005.

\bibitem{dong2020hawq}
Zhen Dong, Zhewei Yao, Daiyaan Arfeen, Amir Gholami, Michael~W Mahoney, and
  Kurt Keutzer.
\newblock Hawq-v2: Hessian aware trace-weighted quantization of neural
  networks.
\newblock {\em Advances in NeurIPS}, 33:18518--18529, 2020.

\bibitem{yang2020bsq}
Huanrui Yang, Lin Duan, Yiran Chen, and Hai Li.
\newblock Bsq: Exploring bit-level sparsity for mixed-precision neural network
  quantization.
\newblock In {\em ICLR}, 2020.

\bibitem{wen2016learning}
Wei Wen, Chunpeng Wu, Yandan Wang, Yiran Chen, and Hai Li.
\newblock Learning structured sparsity in deep neural networks.
\newblock {\em Advances in NeurIPS}, 29, 2016.

\bibitem{molchanov2019importance}
Pavlo Molchanov, Arun Mallya, Stephen Tyree, Iuri Frosio, and Jan Kautz.
\newblock Importance estimation for neural network pruning.
\newblock In {\em Proceedings of the IEEE/CVF Conference on CVPR}, pages
  11264--11272, 2019.

\bibitem{li2016pruning}
Hao Li, Asim Kadav, Igor Durdanovic, Hanan Samet, and Hans~Peter Graf.
\newblock Pruning filters for efficient convnets.
\newblock {\em arXiv preprint arXiv:1608.08710}, 2016.

\bibitem{he2017channel}
Yihui He, Xiangyu Zhang, and Jian Sun.
\newblock Channel pruning for accelerating very deep neural networks.
\newblock In {\em Proceedings of the IEEE international conference on computer
  vision}, pages 1389--1397, 2017.

\bibitem{luo2020neural}
Jian-Hao Luo and Jianxin Wu.
\newblock Neural network pruning with residual-connections and limited-data.
\newblock In {\em Proceedings of the IEEE/CVF Conference on CVPR}, pages
  1458--1467, 2020.

\bibitem{yang2020dverge}
Huanrui Yang, Jingyang Zhang, Hongliang Dong, Nathan Inkawhich, Andrew Gardner,
  Andrew Touchet, Wesley Wilkes, Heath Berry, and Hai Li.
\newblock Dverge: diversifying vulnerabilities for enhanced robust generation
  of ensembles.
\newblock {\em Advances in NeurIPS}, 33:5505--5515, 2020.

\bibitem{laine2016temporal}
Samuli Laine and Timo Aila.
\newblock Temporal ensembling for semi-supervised learning.
\newblock {\em arXiv preprint arXiv:1610.02242}, 2016.

\bibitem{xie2013horizontal}
Jingjing Xie, Bing Xu, and Zhang Chuang.
\newblock Horizontal and vertical ensemble with deep representation for
  classification.
\newblock {\em arXiv preprint arXiv:1306.2759}, 2013.

\bibitem{kilimci2018deep}
Zeynep~Hilal Kilimci and Selim Akyoku{\c{s}}.
\newblock Deep learning-and word embedding-based heterogeneous classifier
  ensembles for text classification.
\newblock {\em Complexity}, 2018.

\bibitem{lee2020layer}
Jaeho Lee, Sejun Park, Sangwoo Mo, Sungsoo Ahn, and Jinwoo Shin.
\newblock Layer-adaptive sparsity for the magnitude-based pruning.
\newblock In {\em ICLR}, 2020.

\bibitem{liu2018rethinking}
Zhuang Liu, Mingjie Sun, Tinghui Zhou, Gao Huang, and Trevor Darrell.
\newblock Rethinking the value of network pruning.
\newblock In {\em ICLR}, 2018.

\bibitem{mocanu2018scalable}
Decebal~Constantin Mocanu, Elena Mocanu, Peter Stone, Phuong~H Nguyen,
  Madeleine Gibescu, and Antonio Liotta.
\newblock Scalable training of artificial neural networks with adaptive sparse
  connectivity inspired by network science.
\newblock {\em Nature communications}, 9(1):1--12, 2018.

\bibitem{frankle2018lottery}
Jonathan Frankle and Michael Carbin.
\newblock The lottery ticket hypothesis: Finding sparse, trainable neural
  networks.
\newblock In {\em ICLR}, 2018.

\bibitem{cai2020rethinking}
Zhaowei Cai and Nuno Vasconcelos.
\newblock Rethinking differentiable search for mixed-precision neural networks.
\newblock In {\em Proceedings of the IEEE/CVF Conference on CVPR}, pages
  2349--2358, 2020.

\bibitem{yang2019deephoyer}
Huanrui Yang, Wei Wen, and Hai Li.
\newblock Deephoyer: Learning sparser neural network with differentiable
  scale-invariant sparsity measures.
\newblock In {\em ICLR}, 2019.

\bibitem{morgan1991experimental}
Nelson Morgan et~al.
\newblock Experimental determination of precision requirements for
  back-propagation training of artificial neural networks.
\newblock In {\em Proc. Second Int’l. Conf. Microelectronics for Neural
  Networks}, pages 9--16. Citeseer, 1991.

\bibitem{wu2016quantized}
Jiaxiang Wu, Cong Leng, Yuhang Wang, Qinghao Hu, and Jian Cheng.
\newblock Quantized convolutional neural networks for mobile devices.
\newblock In {\em Proceedings of the IEEE conference on CVPR}, pages
  4820--4828, 2016.

\bibitem{zhou2017incremental}
Aojun Zhou, Anbang Yao, Yiwen Guo, Lin Xu, and Yurong Chen.
\newblock Incremental network quantization: Towards lossless cnns with
  low-precision weights.
\newblock {\em arXiv preprint arXiv:1702.03044}, 2017.

\bibitem{yu2022hessian}
Shixing Yu, Zhewei Yao, Amir Gholami, Zhen Dong, Sehoon Kim, Michael~W Mahoney,
  and Kurt Keutzer.
\newblock Hessian-aware pruning and optimal neural implant.
\newblock In {\em Proceedings of the IEEE/CVF Winter Conference on Applications
  of Computer Vision}, pages 3880--3891, 2022.

\bibitem{guo2017calibration}
Chuan Guo, Geoff Pleiss, Yu~Sun, and Kilian~Q Weinberger.
\newblock On calibration of modern neural networks.
\newblock In {\em International conference on machine learning}, pages
  1321--1330. PMLR, 2017.

\bibitem{he2016deep}
Kaiming He, Xiangyu Zhang, Shaoqing Ren, and Jian Sun.
\newblock Deep residual learning for image recognition.
\newblock In {\em Proceedings of the IEEE conference on CVPR}, pages 770--778,
  2016.

\bibitem{ron2022experimental}
DA~Ron, PJ~Freire, JE~Prilepsky, M~Kamalian-Kopae, A~Napoli, and SK~Turitsyn.
\newblock Experimental implementation of a neural network optical channel
  equalizer in restricted hardware using pruning and quantization.
\newblock {\em Scientific Reports}, 12(1):8713--8713, 2022.

\bibitem{he2018amc}
Yihui He, Ji~Lin, Zhijian Liu, Hanrui Wang, Li-Jia Li, and Song Han.
\newblock Amc: Automl for model compression and acceleration on mobile devices.
\newblock In {\em Proceedings of ECCV}, pages 784--800, 2018.

\bibitem{zhuang2018discrimination}
Zhuangwei Zhuang, Mingkui Tan, Bohan Zhuang, Jing Liu, Yong Guo, Qingyao Wu,
  Junzhou Huang, and Jinhui Zhu.
\newblock Discrimination-aware channel pruning for deep neural networks.
\newblock {\em Advances in NeurIPS}, 31, 2018.

\bibitem{he2018soft}
Y~He, G~Kang, X~Dong, Y~Fu, and Y~Yang.
\newblock Soft filter pruning for accelerating deep convolutional neural
  networks.
\newblock In {\em IJCAI International Joint Conference on Artificial
  Intelligence}, 2018.

\bibitem{he2019filter}
Yang He, Ping Liu, Ziwei Wang, Zhilan Hu, and Yi~Yang.
\newblock Filter pruning via geometric median for deep convolutional neural
  networks acceleration.
\newblock In {\em Proceedings of the IEEE/CVF conference on CVPR}, pages
  4340--4349, 2019.

\bibitem{peng2019collaborative}
Hanyu Peng, Jiaxiang Wu, Shifeng Chen, and Junzhou Huang.
\newblock Collaborative channel pruning for deep networks.
\newblock In {\em International Conference on Machine Learning}, pages
  5113--5122. PMLR, 2019.

\bibitem{luo2018autopruner}
Jian-Hao Luo and Jianxin Wu.
\newblock Autopruner: An end-to-end trainable filter pruning method for
  efficient deep model inference.
\newblock {\em arXiv preprint arXiv:1805.08941}, 2018.

\bibitem{liu2019metapruning}
Zechun Liu, Haoyuan Mu, Xiangyu Zhang, Zichao Guo, Xin Yang, Kwang-Ting Cheng,
  and Jian Sun.
\newblock Metapruning: Meta learning for automatic neural network channel
  pruning.
\newblock In {\em Proceedings of the IEEE/CVF international conference on
  computer vision}, pages 3296--3305, 2019.

\end{thebibliography}
\end{document}